\documentclass[final]{cvpr}

\usepackage{times}
\usepackage{epsfig}
\usepackage{algorithm}
\usepackage{multicol}
\usepackage{algorithmic}
\usepackage{graphicx}
\usepackage{amsmath}
\usepackage{amssymb}
\usepackage{float}

\usepackage{booktabs}
\usepackage{tabularx}
\usepackage{blindtext}
\usepackage[square,sort,comma,numbers]{natbib}

\usepackage[pagebackref=true,breaklinks=true,colorlinks,bookmarks=false]{hyperref}

\def\cvprPaperID{3831} % *** Enter the CVPR Paper ID here
\def\confYear{CVPR 2021}

\begin{document}

%%%%%%%%% TITLE
\title{Removing Class Imbalance using \\ Polarity-GAN: An Uncertainty Sampling Approach}

\author{Kumari Deepshikha\\
NVIDIA\\
{\tt\small dkumari@nvidia.com}
% For a paper whose authors are all at the same institution,
% omit the following lines up until the closing ``}''.
% Additional authors and addresses can be added with ``\and'',
% just like the second author.
% To save space, use either the email address or home page, not both
\and
Anugunj Naman\\
IIIT  Guwahati\\
{\tt\small anugunjjha@gmail.com}
}

\maketitle

%%%%%%%%% ABSTRACT
\begin{abstract}
   Class imbalance is a challenging issue in practical classification problems for deep learning models as well as for traditional models. Traditionally successful countermeasures such as synthetic over-sampling have had limited success with complex, structured data handled by deep learning models. In this work, we propose to use a Generative Adversarial Network (GAN) equipped with a generator network G, a discriminator network D and a classifier network C to remove the class-imbalance in visual data sets. The generator network is initialized with auto-encoder to make it stable. The discriminator D ensures that G adheres to class distribution of imbalanced class.

In conventional methods, where Generator G competes with discriminator D in a min-max game, we propose to further add an additional classifier network to the original network. Now, the generator network tries to compete in a min-max game with  Discriminator as well as the new classifier that we have introduced. An additional condition is enforced on generator network G to produce points in the convex hull of desired imbalanced class. Further the contention of adversarial game with classifier C, pushes conditional distribution learned by G towards the periphery of the respective class, compensating the problem of class imbalance. Experimental evidence shows that this initialization results in stable training of the network. We achieve state of the art performance on extreme visual classification task on the FashionMNIST, MNIST, SVHN, ExDark, MVTec Anomaly Detection dataset, Chest X-Ray dataset and others.

\end{abstract}

\section{Introduction}

The problem of class imbalance occurs when the distribution of samples among all possible classes is uneven, which results in classes with widely varying sample sizes. Here, the distribution of examples across the known categories is biased or skewed. The distribution can vary from a slight bias to a severe imbalance where there is one example in the minority class for hundreds\cite{chawla2002smote}, thousands, or millions of illustrations in the majority class or classes. Datasets with class imbalance pose a severe challenge for predictive modeling as most of the machine learning algorithms used for classification are designed around the assumption of an approximately equal number\cite{kotsiantis2006handling}\cite{nguyen2009learning}\cite{japkowicz2002class} of examples for each class. Hence, for the same reason, machine learning models obtained after training on data sets with severe class imbalance often have poor predictive performance, specifically for the minority class (a class with too few training samples). 

Various methods have been proposed in past to handle class imbalance\cite{ramentol2012smote}\cite{lee2012anomaly}\cite{suh2019generative}\cite{vuttipittayamongkol2020improved}\cite{lu2015feature}\cite{gauthier2014conditional}\cite{odena2017conditional}\cite{douzas2018effective}. Most popular of these approaches can be classified into two major categories (i) Data-level solution \cite{johnson2019survey}, and (ii) Algorithmic level solution. Data-Level methods have few initial success stories and are often based on under-sampling, oversampling, and synthetic sampling. These approaches face certain drawbacks. While performing under sampling~\cite{yen2006under}, there is a high risk of removing the data point that may contain important information about the predictive class. On the other hand, oversampling may result in overfitting to the training data.  Synthetic sampling(SMOTE)\cite{chawla2002smote} based approach synthetically manufacture observations of unbalanced classes similar to the existing one using nearest neighbor classification. However, it is not clear how one synthetically manufacture observation of an unbalanced class when the number of observations is infrequent. For example, we may have only one picture of a rare species we would like to identify using an image classification algorithm. Algorithm-Level methods include cost-sensitive\cite{thai2010cost} and hybrid approaches. At the Algorithm-Level\cite{li2018cost}, there are a variety of methods that have been used. Such as hybrid/ensemble approaches includes a Bayesian Optimization Algorithm that maximizes Matthew's Correlation Coefficient by learning optimal weights for the positive and negative classes \cite{snoek2012practical} and an approach that combines Random Over-Sampling (ROS) and Support Vector Machines (SVMs)\cite{mathew2017classification}.

Recently, generative adversarial networks (GANs)\cite{goodfellow2014generative} have emerged as a class of generative models approximating the real data distribution. GANs have been widely used for the task of image generation\cite{yang2017lr}, image super-resolution\cite{ledig2017photo} and semi-supervised learning\cite{springenberg2015unsupervised}. State of the art performances for these tasks is obtained by utilizing generative adversarial networks' power to generate samples from the real data distribution. Various work including Conditional GANs (CGAN) \cite{gauthier2014conditional} and auxiliary classifier GANs (ACGAN) \cite{odena2017conditional}   further extend GANs by conditioning the training procedure on the class labels for the classifier. It was demonstrated in \cite{douzas2018effective} that data generated by conditioned GAN improved the classification performance. However, to the best of our knowledge, there has not been any substantial work done for extreme visual classification tasks while using a GAN to generate samples. The main hurdle of using GAN for extreme visual classification tasks comes from the fact that it usually requires many samples to train a generative adversarial network. More often than not, too few training samples are available for most of the classes in datasets with class imbalance.  Apart from that, the training process's instability remains a major practical challenge while training a GAN.  Also, conditional GANs only performs well in the case where class boundaries are clear.

When a supervised model makes a prediction, it often provides a confidence score for that prediction. In this paper, we use modified GAN as an uncertainty sampling method. Getting feedback when the model is uncertain about prediction has resemblance with active learning popularly known in the literature as uncertainty sampling for active learning\cite{yang2015multi}. Using uncertainty-based sampling (Manali and Mustafa et al.) has shown that distinguishing between different types of uncertainties has a drastic impact on the learning efficiency\cite{sharma2017evidence}. This paper uses a modified Generative Adversarial Network for uncertainty sampling, which we further utilize to sample hard examples and use it for extreme visual classification. To achieve the goal,  we also propose adding a second classifier network to the original network.  Here, the Generator network tries to compete in a min-max game with  Discriminator and the classifier we introduced in the GAN module. An additional condition is enforced on generator network G to produce points within the desired imbalanced class convex hull. The generator network is initialized with auto-encoder to make it stable. The discriminator D ensures that G adheres to the class distribution of imbalanced class. 
We make the following contributions to the paper.  

\begin{itemize}
    \item Changes introduced in the GAN, i.e., generating points in the convex hull while keeping it in contention with classifier make sure that the points generated are on the boundary, which helps classifier find a more deterministic decision boundary. 
    \item  Using autoencoders and generating points in convex hulls provide better convergence and prevent points from mixing into different classes, respectively.
    \item Using uncertainty help in making the distribution more even. We have done testing extensively on real-world data sets having a high imbalance class, which is not performed in previous methods.
\end{itemize}

 We used the below parameters to measure uncertainty sampling. 
\begin{itemize}
\item  Least Confidence: the difference between the most confident prediction and 100$\%$ confidence.
\item Margin of Confidence: the difference between the top two most confident predictions.
\item Ratio of Confidence: the ratio between the top two most confident predictions.
\item Entropy: the difference between all predictions, as defined by information theory with our methods.
\end{itemize}

We achieve state of the art performance on extreme visual classification tasks on the FashionMNIST\cite{xiao2017fashion}, CIFAR10\cite{krizhevsky2009learning} MNIST\cite{lecun1998mnist}, SVHN\cite{netzer2019street}, ExDark\cite{loh2019getting}, UCF101\cite{soomro2012ucf101}, Anomaly Detection, detection\cite{bergmann2019mvtec}, COVID datasets\cite{wang2020covid}. In the rest of the paper, we will discuss Methodology used, Algorithms, Results and Related work. our method lead an improvement of $5\%$ for class with least sample i.e. 1:10 ratio in CIFAR-10 dataset.

%------------------------------------------------------------------------

\section{Related Work}
Several works have been done to solve the imbalanced data problem in Machine Learning/Deep learning\cite{ramentol2012smote}\cite{lee2012anomaly}\cite{suh2019generative}\cite{vuttipittayamongkol2020improved}\cite{lu2015feature}\cite{gauthier2014conditional}\cite{odena2017conditional}\cite{douzas2018effective}. Huang et al.\cite{huang2016learning} handled imbalanced-data by integrating the sampling process into Convolution Neural Network (CNN) \cite{srivastava2014dropout}. The sampling process was done by dividing the data into several clusters, then a pair of databases on the determined condition was retrieved. These pairs of data were then inputted to CNN. After this, the process was then repeated from clustering until the model reaches the convergences. Yan et al.\cite{yan2015deep}, and Khan et al.\cite{khan2017cost}handled imbalanced data by modifying a CNN. Wei et al.\cite{wei2015deep} handled an imbalance in object recognition by designing an image-cutting technique used as additional training data. George et al.\cite{george2015image} and Mostajabi et al.\cite{mostajabi2015feedforward} used cost-sensitive to handle imbalanced data in scene-parsing. The techniques used in these researches were specifically designed for the data set that was used in their research. Thus, it isn't easy to implement techniques for different data sets. These techniques also had other limitations addressed by Aida Ali et al.\cite{ali2015classification}.

Recently, generative adversarial networks (GANs) (Goodfellow et al., 2014, Schmidhuber, 2020) \cite{goodfellow2014generative}\cite{schmidhuber2020generative} have emerged as a class of generative models approximating the real data distribution. GANs represent a class of generative models based on a game-theoretic framework in which a generator network G competes against an adversary  D, a discriminator. GANs aim to approximate the probability distribution function that certain data is assumed to be drawn from. 
Although GAN has shown success in realistic image generation, the training is not easy; The process is known to be unstable and slow.

Later on, Arjovsky, Chintala et al. had introduced WGAN (Wasserstein GAN)\cite{arjovsky2017wasserstein} (Figure \ref{fig:GAN}) to solve the stability problem. In this work, the Wasserstein-1 distance, also known as the earthmover (EM) distance, is used as a cost function, a continuous and more appropriate metric for measuring the distance between two distributions. The EM distance does not suffer from vanishing gradients; by contrast, the JS (Jensen–Shannon divergence)\cite{menendez1997jensen} divergence in the GAN does not supply useful gradients to the generator.

\textbf{Conditional GANs (CGAN):} (Mirza, Osindero, 2014\cite{mirza2014conditional} (Figure \ref{fig:CGAN} in the Appendix). This model's novelty is that they implemented classification shared weights with discriminate models and forced D models to concentrate on small classes' features. They simultaneously train a generative model and a fine-grained classifier. 

Later, Giovanni Mariani, Florian Scheidegger et al. \cite{mariani2018bagan}had introduced applied class conditioning in the latent space to drive the generation process towards a target class. They called it BAGAN.
Auxiliary Classifier GANs (ACGAN) (Odena, Olah, Shlens, 2017) (Figure \ref{fig:ACGAN} in the Appendix) is a class-conditional extension of GANs and offers a simple method for providing varying amounts of control in the image generation process by adding a classification layer to the discriminator. The two generator objectives, i.e., classifying real/fake and classifying the label, as shown in equation \ref{eq:3} and \ref{eq:4}, are contradictory. The generator here is made to generate an image that is real or generate an image for minority labels, which causes deterioration of images generated. Thus the quality of the data generated is incomparable with that of the real training data.

Douzas and Bacao (2018)\cite{douzas2018effective} demonstrate that the data generated by CGAN improves the classification's performance. However, GAN and conditional GANs, such as CGAN and ACGAN, have limitations. The instability of the training process remains these GANs as a challenge in practice. The other problem is the influence of noise. As the generator of the GANs takes as input a random noise vector and outputs a synthetic image, we need the classifier for the conditional GANs to consider the influence of noise. Also, conditional GANs perform well in the hypothesis that the class boundaries are clear. In a real-world dataset, this hypothesis does not hold because the boundaries between classes are often unclear and ambiguous.

There is another approach taken by Suh, Sungho et al. in CE-GAN\cite{suh2020cegan}, also known as classification enhancement GAN. In this approach, firstly, the class label is classified by an independent classifier. Step 2 involves training a classifier with the real and augmented dataset to verify the classification performance's improvement. In this approach, data augmentation is done to increase samples.  It is entirely unclear whether the performance improvement is because of progress in the performance for minority class or overall improvement in the classification performance.

We found missing major components in all GAN-based approaches that the authors have not performed any experimental evaluation on real-world data set with imbalance distribution. Such as like anomaly detection\cite{bergmann2019mvtec}, Healthcare data, fault diagnosis, face recognition, and COVID datasets \cite{wang2020covid}, medical diagnosis, detection of oil spillage in satellite images. Majorly all experiments have been performed with CIFAR-10, MNIST data sets. We find out that state-of-the-art architecture like WGAN tends to overfit those datasets for minority classes.

\section{Background}
Before going into details of our method, we discuss the background on generative adversarial networks, which we build upon and use for our baseline. For the baseline, the image generated using GANs is further fed into the classifier to learn the prediction function. 
\subsection{GAN}
It consist of two major components (i) a \textit{Discriminator} D and (ii) a \textit{Generator} G .
The discriminator estimates the probability of a given sample coming from the real dataset. It works as a critic and is optimized to tell the fake samples from the real ones.

\begin{figure}[ht]
\begin{center}
\includegraphics[width=8.0cm]{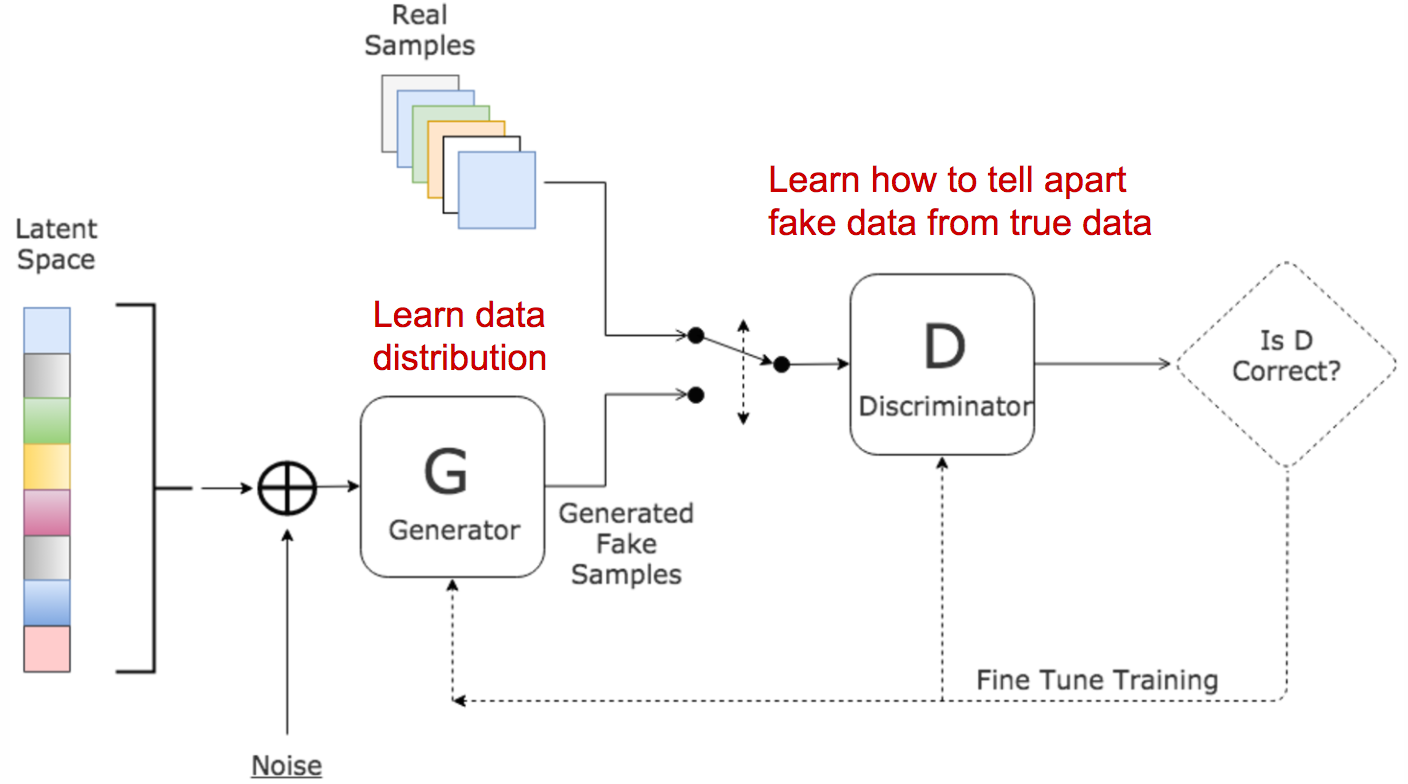}    % The printed column width is 8.4 cm.
\caption{Architecture of a generative adversarial network, Source: \cite{ganmanga}}
\label{fig:GAN}
\end{center}
\end{figure}

The generator G outputs synthetic samples given a noise variable input z (z brings in potential output diversity). It is trained to capture the real data distribution so that its generative examples can be as accurate as possible. In other words, it can trick the discriminator into offering a high probability.
The objective function of the min-max game between the generator and the discriminator is expressed as follows:

\begin{equation} \label{eq:1}
\begin{split}
\min_{G} \max_{D} V(D,G) &=  \mathbb{E}_{x \sim p_{data}(x)}[\log D(x)]  \\
&~~ +\mathbb{E}_{z \sim p_{z}(z)}[\log (1 - D(G(z)))]
\end{split}
\end{equation}
where ${V(D,G)}$ is the objective function, ${G}$ and ${D}$ denote the generator and discriminator respectively.
Here figure \ref{fig:GAN} shows architecture of GAN. In this both discriminator and Generator competes with each other in Min-Max Game.

\subsection{WGAN}
The architecture design of WGAN is almost the same as that of vanilla GAN discussed previously. However,   the critic does not have a non-linearity (sigmoid) on the output. The significant difference is only on the cost function used to measure the discrepancy between two distributions. WGAN uses Wasserstein-1 distance to measure the disparity between real data distribution and generated data distribution. Due to Kantorovich-Rubinstein's duality, the training becomes tractable for WGAN. The architectural representation is given in Figure \ref{fig:GAN}. 
The objective function of WGAN is:

\begin{equation} \label{eq:5}
\begin{split}
V(D,G) = \min_{G} \max_{D \in{\mathcal{D}}} \mathbb{E}_{x \sim p_{r}}[D(x)] - \mathbb{E}_{x \sim p_{g}}[D(\tilde{x})]
\end{split}
\end{equation}

Here, in equation \ref{eq:5}, $\mathcal{D}$ denotes the set of 1-Lipschitz functions (A Lipschitz function is a function ${f}$ such that ${—f(x)-f(y)— \leq K—x-y—}$ for all x and y, where ${K}$ is a constant independent of ${x}$ and ${y}$), meaning discriminator loss should follow the Lipschitz constraint.

\subsection{BAGAN}
\begin{figure}[ht]
\begin{center}
\includegraphics[width=8.0cm]{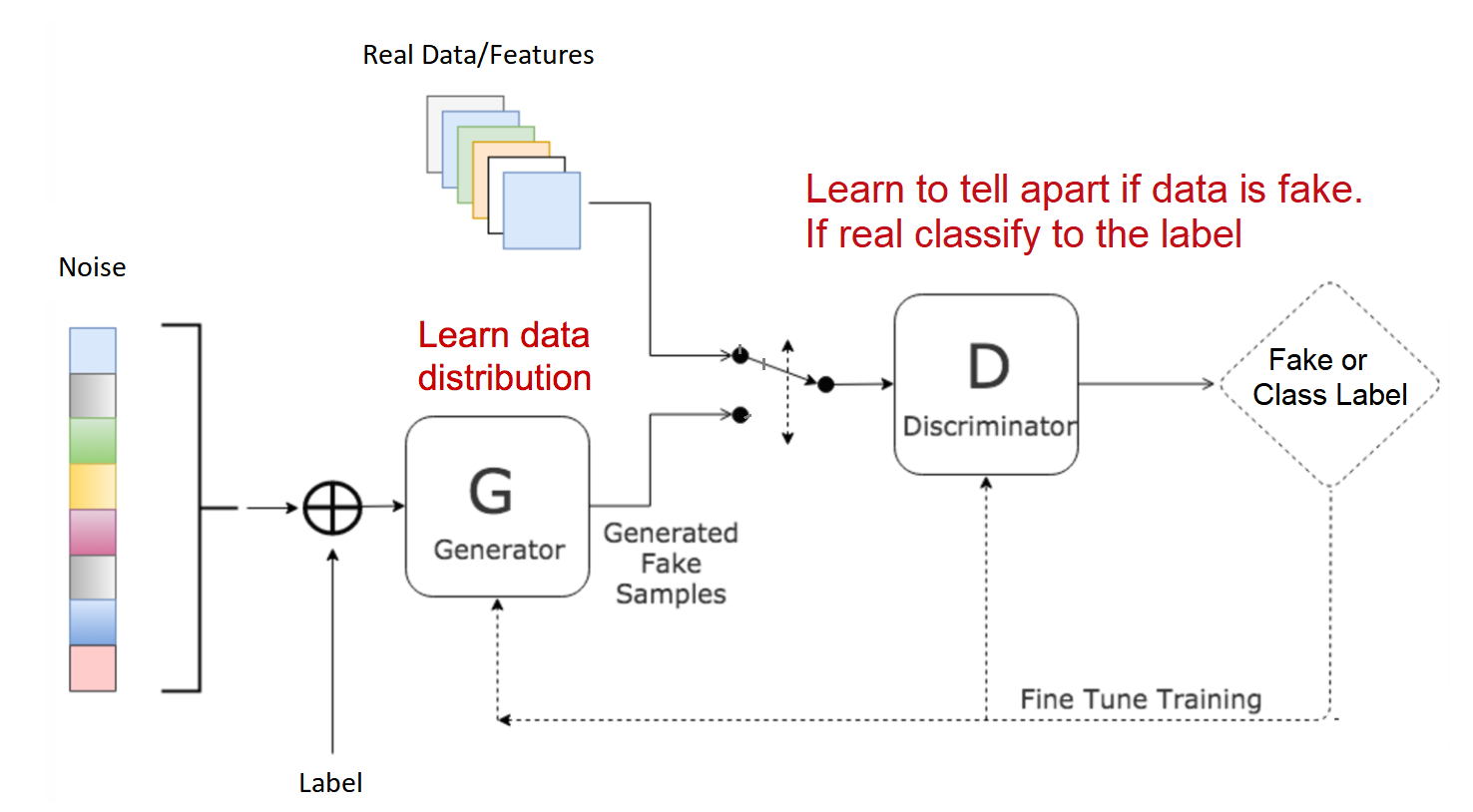}    % The printed column width is 8.4 cm.
\caption{{Framework of Balancing GAN (BAGAN).}} 
\label{fig:BAGAN}
\end{center}
\end{figure}

Balancing GAN (BAGAN) uses as an augmentation tool to restore balance in imbalanced data sets. The GAN generator is initialized with an auto-encoder module that enables us to learn an accurate class-conditioning in the latent space. Here in figure \ref{fig:BAGAN} shows the architectural details of BAGAN.

\section{Proposed Method}
In the previous section, we discussed the relevant background. In this section, we will discuss our proposed approach. But, let us first discuss the major challenges of GANs while dealing with imbalanced datasets. Suppose we train a vanilla GAN on all available data and generate many random samples. The generated samples are mostly from the majority class due to the more number of training samples. However, due to the small number of available samples for minority classes, the training procedure makes our model memorize the samples and leads to overfitting.

We propose a variant of a generator that solves the above problem by generating samples in the training samples' convex hull for each class. The generator here competes in a min-max game with a classifier to generate samples difficult for the classifier to classify. In contrast, the classifier pushes the generator to generate more challenging samples, eventually causing the generator to generate the samples near the training samples' convex hull. It helps the classifier to find a more deterministic decision boundary. The generator also competes in a min-max game with a discriminator to ensure the generator's samples are within each label's class distribution and does not suffer from leakage into other classes.

\begin{figure}[ht]
\begin{center}
\includegraphics[width=5.0cm]{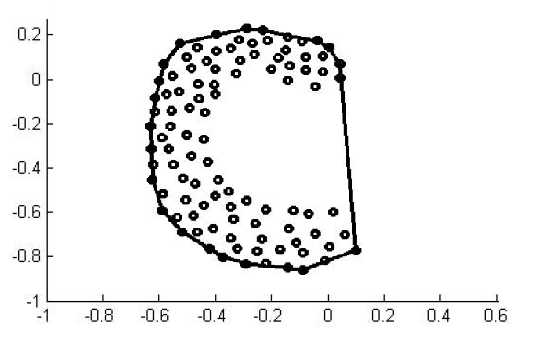}    % The printed column width is 8.4 cm.
\caption{{Convex hull can not represent exactly the borders of all sets of points.}} 
\label{fig:hull1}
\end{center}
\end{figure}

\begin{figure}[ht]
\begin{center}
\includegraphics[width=5.0cm]{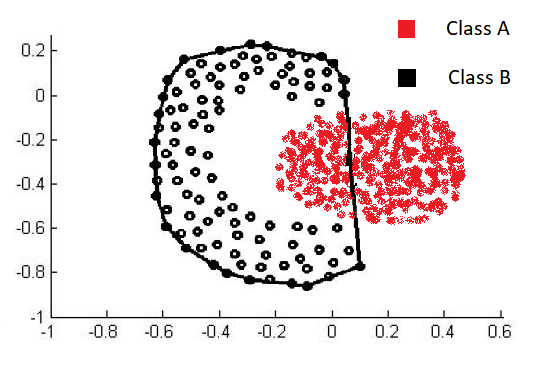}    % The printed column width is 8.4 cm.
\caption{{Leaking of generated points in other label's class distribution.}} 
\label{fig:hull3}
\end{center}
\end{figure}

\textbf{Leakage of Points}: The convex hull is defined as a set of points is the minimum convex set containing all points of the set. The major problem with the convex hull is that in many cases, it can not represent the shape of a set, i.e., for a set of points having interior "corners" or concavities. It omits the points that determine the border of those areas. \cite{chau2011border}. An example can be seen in Figure \ref{fig:hull1}. In such cases, if another class has training points inside the convex hull as shown in Figure \ref{fig:hull3}, then the generator will generate points in the intersecting region of both classes.  To make it difficult for the classifier to classify the points, it diverts us from our main motive to generate points near the boundary. Therefore, we use discriminator here as a check and balance to make sure the generator adheres to class distribution.

\begin{algorithm}[h!]
\caption{Polarity GAN Algorithm}\label{algo:1}
\begin{algorithmic}[1]
\REQUIRE ${G}$: Generator, ${D}$: Discriminator, ${C}$: Classifier, ${X}$: Training Data, ${Y}$: Labels of Training Data ${X}$.
${F}$ Feature Extractor, ${z}$: noise
\ENSURE
\WHILE{ not converged }
\STATE Update weights of feature extractor ${F}$ by gradient descent on ${(C(F(X)), Y)}$ while keeping weights of classifier ${C}$ fixed.
\FOR{j steps}
\STATE Update classifier ${C}$ and discriminator $D$ on training data ${X}$ by gradient descent on ${(C(F(X)), Y)}$ and ${(D(F(X)), 1)}$, while keeping ${F}$ fixed.
\STATE Sample data ${X^{'}}$ $\in$ ${z}$.
\STATE Generate labels ${Y^{'}}$ for ${X^{'}}$ into one of the ${k-1}$ minority classes with probability $\propto{(P_{k} - P_{i})}$ $\forall$ $i \in K \setminus k$ and ${K}$ set of class labels.
\STATE Update classifier ${C}$ and discriminator $D$ on ${X^{'}}$ by gradient descent on ${(C(G(X^{'}|Y^{'})), Y^{'})}$ and ${(D(G(X^{'}|Y^{'})), 1)}$ keeping generator ${G}$ fixed.
\STATE Sample data ${X^{''}}$ $\in$ $z$.
\STATE Generate labels ${Y^{''}}$ for ${X^{''}}$ into one of the ${k-1}$ minority classes with  equal probability
\STATE Update generator ${G}$ by gradient on ${(C(G(X^{''}|Y^{''})), Y^{''})}$ keeping  ${C}$ fixed.
\STATE Update generator${G}$ by gradient on ${(D(G(X^{''}|Y^{''})), 1)}$ keeping ${D}$ fixed.
\ENDFOR
\ENDWHILE
\end{algorithmic}
\end{algorithm}

The generator network consists of two mapping function: the first mapping ${Q}$ from noise ${z}$ conditioned on class label ${k}$ to a pseudo space where it is combined with original training samples to create another mapping P from pseudo space to the convex hull of the training samples. Taking inspiration from WGAN we have introduced Wasserstein distance in our cost function.  Wasserstein distance is a continuous and more appropriate metric for measuring the distance between two distributions. The EM distance does not suffer from vanishing gradients; by contrast, the GAN's JS divergence does not supply useful gradients to the generator. The generator function can be defined as:
\begin{equation} \label{eq:6}
\begin{split}
G(z|k_{i}) = \sum_{j=1}^{n_{k}} P(Q(z|k_{i}))x_{j}
\end{split}
\end{equation}
In equation \ref{eq:6}, ${x_{j}}$ represents a single training sample belonging to the  class label ${k}$ and ${n_{k}}$ represents number of training samples in label ${k}$ with condition ${\sum_{j=1}^{n_{k}} P(Q(z|k_{i})) = 1}$ using softmax activation.

We have also applied a pragmatic use of auto-encoders inspired by BAGAN to initialize the GAN close to the right solution and far from mode collapse. The decoder of the auto-encoder matches the topology of the generator network. 
\paragraph{Autoencoder training}: The auto-encoder is trained by using all the images in the training dataset. The auto-encoder has no explicit class knowledge; it processes all images from the majority and minority classes unconditionally.

\begin{figure}[ht]
\begin{center}
\includegraphics[width=8.0cm]{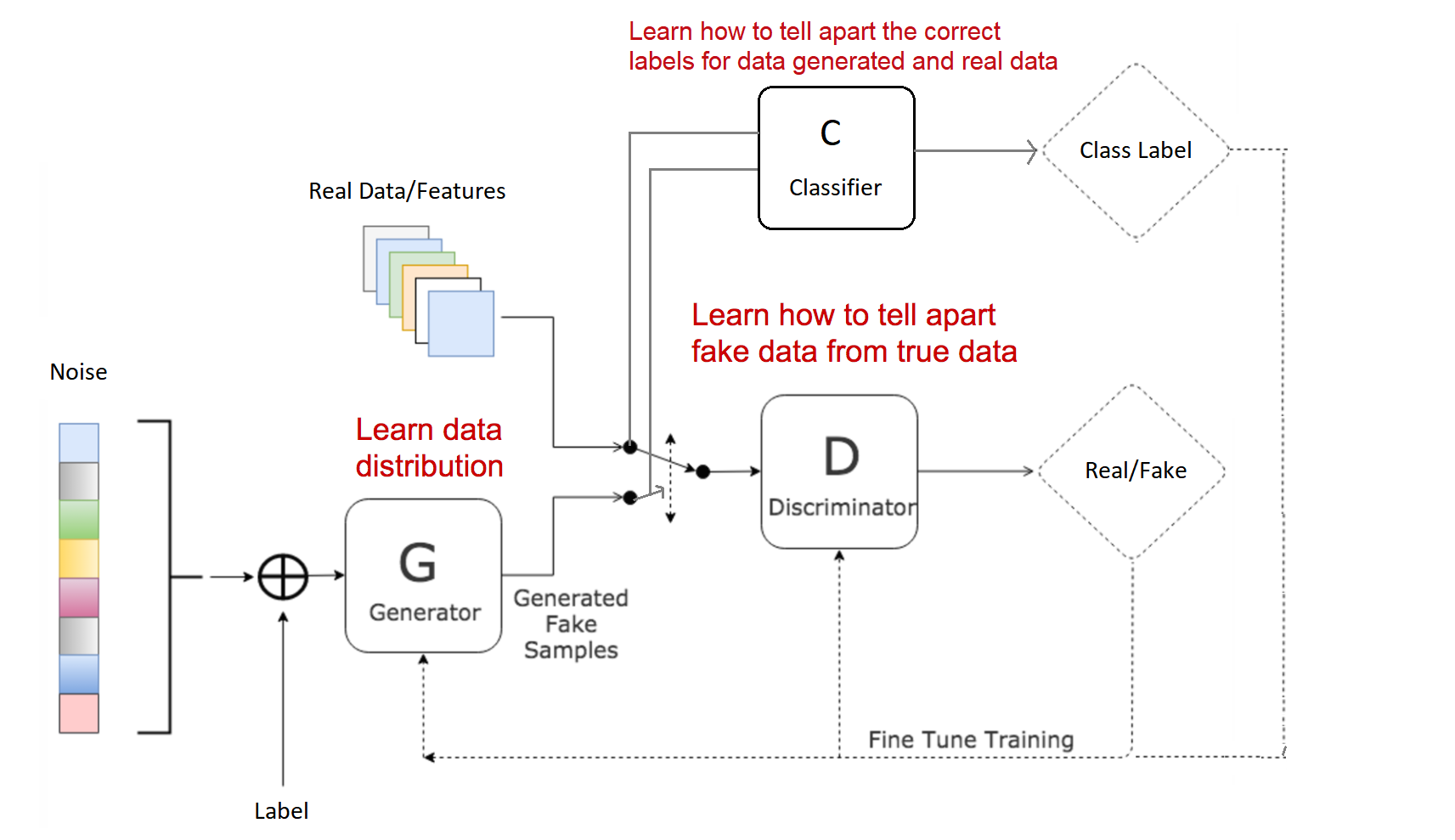}    % The printed column width is 8.4 cm.
\caption{{Frame work of Polarity-GAN.}} 
\label{fig:PGAN}
\end{center}
\end{figure}

The complete framework of our method is described in Figure \ref{fig:PGAN}. The training details of P-GAN are summarized in Algorithm  \ref{algo:1}. Assuming without loss of generality, the prior probability of ${i_{th}}$ label be ${p_{i}}$ then class can be ordered in the following manner ${p_{1}} \le {p_{2}} ... \le {p_{k}}$. Then final objective function can be written as follows:

{\small
\begin{align} \label{eq:7}
\begin{split}
\min_{G} \max_{D} \max_{C} V(G,D, C) =  \sum_{i \in {D, C}} \mathcal{S}_{i}
\end{split}
\end{align}

\begin{equation} \label{eq:8}
\begin{split}
\mathcal{S}_{D} &= p_{i}\mathbb{E}_{x \sim p_{data}(x)}[D(x|i)]\quad  \\ 
&\qquad \qquad -(p_{k} - p_{i})\mathbb{E}_{z \sim p_{z}(z)}[(D(G(z|i)))]
\end{split}
\end{equation}
{\small
\begin{equation} \label{eq:9}
\begin{split}
&\mathcal{S}_{C} = p_{i}\mathbb{E}_{x \sim p_{data}(x)}[\log C(x|i)]\quad  \\ 
&\quad +\sum_{j \in K \setminus \{i\}}\quad (p_{k} - p_{j})\mathbb{E}_{x \sim p_{data}(x)}[\log 1 - C(x|j)]\quad  \\ 
&\qquad \qquad  +p_{i}\mathbb{E}_{z \sim p_{z}(z)}[\log (C(G(z|i)))]\quad  \\  
& +\sum_{j \in K \setminus \{i\}} (p_{k} - p_{j})\mathbb{E}_{z \sim p_{z}(z)}[\log (1 - C(G(z|j)))]\quad
\end{split}
\end{equation}
}
}
In the equation \ref{eq:8} and \ref{eq:9}, $\mathcal{S}_{D}$ and $\mathcal{S}_{C}$ represent the objective function of the min-max game between generator and discriminator and generator and classifier respectively. The generator's objective function is based on Wasserstein's loss, while for classifier and generator, it is based on cross-entropy loss.  We also have augmented the data using data augmentation methods to avoid the under-fitting scenario.

\section{Experiments}
Now that we have discussed our proposed approach in the previous section, in this section, we experimentally evaluate our proposed method's performance on real-world datasets and compare it with other baselines. 

\begin{table*}[h]
  \centering
  \begin{tabular}{ |c|c|c|c|c|c|c|c| } 
\hline

\textbf{Dataset} & \textbf{SMOTE-SVM} & \textbf{Resnet+RF} & \textbf{Resnet50} & \textbf{AC-GAN} & \textbf{BAGAN} & \textbf{WGAN-GP} & \textbf{P-GAN (Ours)}\\\hline
MNIST & 0.82 & 0.83 & 0.86 & 0.89 & \textbf{0.91} & 0.90 & \textbf{0.91} \\
Fashion-MNIST & 0.69 & 0.77 & 0.81 & 0.83 & 0.85 & 0.83 & \textbf{0.90} \\
CIFAR10 & 0.26 & 0.33 & 0.68 & 0.72 & 0.73 & 0.72 & \textbf{0.75} \\
SVHN & 0.58 & 0.56 & 0.73 & 0.74 & 0.75 & 0.74 & \textbf{0.78} \\
UCF101 & 0.19 & 0.29 & 0.65 & 0.66 & 0.69 & 0.65 & \textbf{0.74} \\
ExDark & 0.20 & 0.27 & 0.51 & 0.51 & 0.52 & 0.52 & \textbf{0.54} \\
Covidx & 0.45 & 0.55 & 0.89 & 0.92 & 0.94 & 0.92 & \textbf{0.96} \\
Kaggle Birds & 0.47 & 0.53 & 0.95 & 0.94 & 0.95 & 0.96 & \textbf{0.98} \\
Cloud  & 0.32 & 0.39 & 0.63 & 0.64 & 0.66 & 0.63 & \textbf{0.69}\\
\hline
\end{tabular}
\vspace{1mm}
  \caption{F1-Score of different state of ART models on different data sets.}
  \label{tab:1}
\end{table*}

{\small
\begin{table*}[h]
  \centering
  \begin{tabular}{ |c|c|c|c|c|c|c|c| } 
\hline

\textbf{Dataset} & \textbf{SMOTE-SVM} & \textbf{Resnet+RF} & \textbf{Resnet50} & \textbf{AC-GAN} & \textbf{BAGAN} & \textbf{WGAN-GP} & \textbf{P-GAN (Ours)}\\\hline
MNIST & 0.80 & 0.87 & 0.91 & 0.92 & \textbf{0.94} & 0.92 & 0.93\\
Fashion-MNIST & 0.66 & 0.70 & 0.80 & 0.82 & 0.84 & 0.82 & \textbf{0.90} \\
CIFAR10 & 0.23 & 0.44 & 0.69 & 0.70 & 0.72 & 0.71 & \textbf{0.77} \\
SVHN & 0.56 & 0.60 & 0.65 & 0.67 & 0.70 & 0.68 & \textbf{0.74} \\
UCF101 & 0.19 & 0.32 & 0.74 & 0.75 & 0.77 & 0.75 & \textbf{0.79} \\
ExDark & 0.21 & 0.30 & 0.47 & 0.49 & 0.51 & 0.51 & \textbf{0.53} \\
Covidx & 0.48 & 0.53 & 0.90 & 0.91 & 0.93 & 0.92 & \textbf{0.96} \\
Kaggle Birds & 0.47 & 0.51 & 0.95 & 0.95 & 0.96 & 0.96 & \textbf{0.98} \\
Cloud  & 0.34 & 0.37 & 0.64 & 0.66 & 0.67 & 0.65 & \textbf{0.70}\\
\hline
\end{tabular}
\vspace{1mm}
  \caption{Average Accuracy/(True Positive+True Negative) of different state of ART models on different data sets.}
  \label{tab:2}
\end{table*}
}

{\small
\begin{table*}[h]
  \centering
  \begin{tabular}{ |c|c|c|c|c|c|c|c|c| } 
\hline
\textbf{Dataset} & \multicolumn{2}{c|}{\textbf{Least Confidence}} & \multicolumn{2}{c|}{\textbf{Margin of Confidence}} & \multicolumn{2}{c|}{\textbf{Ratio of Confidence}} & \multicolumn{2}{c|}{\textbf{Entropy}}\\\hline

- & Previous Best & Ours & Previous Best & Ours & Previous Best & Ours & Previous Best & Ours\\\hline
MNIST & \textbf{0.07} & 0.08 & \textbf{0.15} & 0.16 & \textbf{0.09} & 0.08 & 0.16 & \textbf{0.15}\\
Fashion MNIST & 0.14 & \textbf{0.12} & 0.19 & \textbf{0.19} & \textbf{0.05} & 0.08 & \textbf{0.14} & 0.16\\
CIFAR10 & 0.62 & \textbf{0.56} & 0.76 & \textbf{0.72} & 0.45 & \textbf{0.41} & 0.52 & \textbf{0.50}\\
SVHN & 0.26 & \textbf{0.24} & 0.33 & \textbf{0.32} & 0.13 & \textbf{0.11} & 0.52 & \textbf{0.29}\\
UCF101 & \textbf{0.63} & 0.64 & 0.87 & \textbf{0.86} & 0.69 & \textbf{0.66} & 0.66 & \textbf{0.61}\\
ExDark & 0.58 & \textbf{0.56} & 0.92 & \textbf{0.92} & 0.86 & \textbf{0.85} & 0.40 & \textbf{0.39}\\
Covidx  & 0.12 & \textbf{0.07} & 0.18 & \textbf{0.14} & 0.13 & \textbf{0.10} & 0.15 & \textbf{0.11}\\
Kaggle Birds  & 0.05 & \textbf{0.02} & 0.06 & \textbf{0.03} & 0.03 & \textbf{0.01} & 0.07 & \textbf{0.05}\\
Cloud  & 0.68 & \textbf{0.62} & 0.72 & \textbf{0.72} & \textbf{0.51} & 0.61 & 0.52 & \textbf{0.48}\\
\hline
\end{tabular}
\vspace{2mm}
  \caption{Uncertainty Sampling.}
  \label{tab:3}
\end{table*}
}

\subsection{Datasets and Model Details}

For the evaluation of proposed methods we have used 9 Datasets. These data-sets are: MNIST\cite{lecun1998mnist}, fashion-MNIST\cite{xiao2017fashion}, CIFAR-10\cite{krizhevsky2009learning}, SVHN\cite{goodfellow2013multi}, ExDark\cite{loh2019getting}, UCF101\cite{soomro2012ucf101},COVID datasets\cite{cohen2020covid}, Kaggle Bird dataset\footnote{\url{https://www.kaggle.com/gpiosenka/100-bird-species}} and Satellite Cloud Anomaly detection\footnote{\url{https://www.kaggle.com/ashoksrinivas/cloud-anomaly-detection-images}} .

\begin{figure}[h!]
\includegraphics [width=84mm]{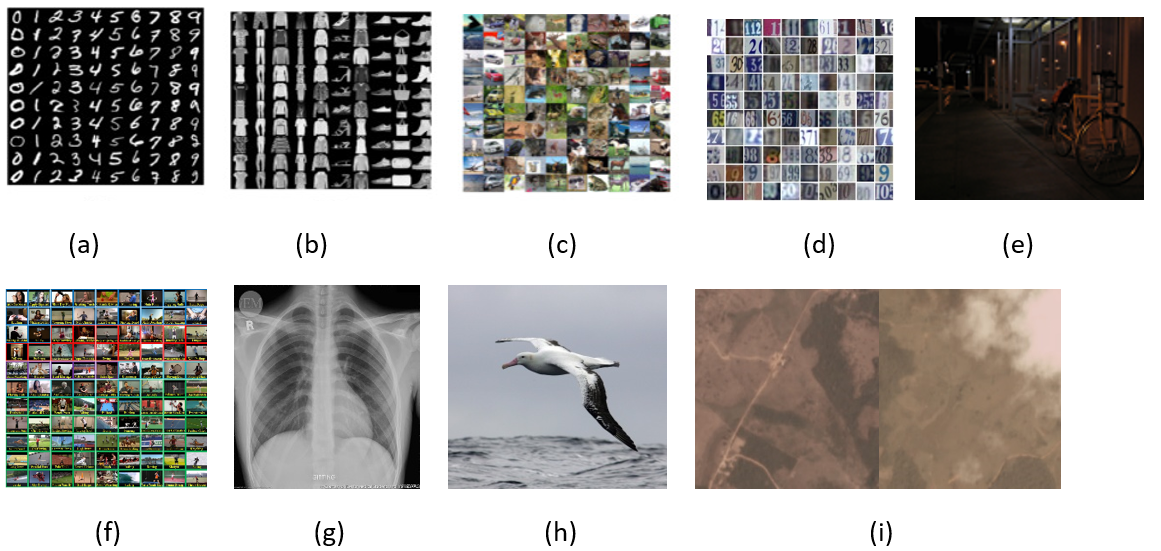}
\caption{Example images of each dataset:  (a) MNIST, (b) Fashion-MNIST,(c)  CIFAR-10, (d) SVHN, (e) Exdark, (f) UCF101, (g) CovidX (h) Kaggle Birds (i) Cloud Anomaly}
\label{fig:Datsets}
\end{figure}

Example images from each dataset are shown in Figure \ref{fig:Datsets}

We have taken Resnet-50 as a base model for the extraction of features with weights of imagenet. The hyperparameters for generator, discriminator, and classifier are fixed using grid search. The optimizer used is Adam Optimizer, with the learning rate set at 0.0002 and exponential decay rate for the 1st-moment estimates at 0.6.

\subsection{Experimental Settings and Results}
\subsubsection{MNIST}
The MNIST database of handwritten digits\cite{lecun1998mnist} has a training set of 60,000 examples, and a test set of 10,000 examples. The digits have been size-normalized and centered in a fixed-size image.  We have used all ten classes in the experiment with examples in following ratio: [5000, 4000, 3000, 2000, 1500, 1000, 500, 250, 100, 50]. The data points selected were at random. The overall F1-score of PGAN is 0.91. 

\subsubsection{Fashion MNIST}
\begin{figure}[!htb]
\minipage{0.23\textwidth}
  \includegraphics[width=\linewidth]{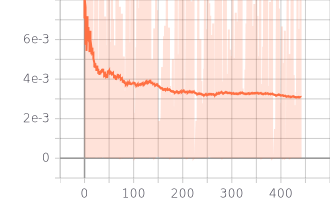}    % The printed column width is 8.4 cm.
 \caption{{Training loss Vs Iteration with our model}} 
\label{fig:Loss Vs Iteration}
\endminipage\hfill
\minipage{0.23\textwidth}
  \includegraphics[width=\linewidth]{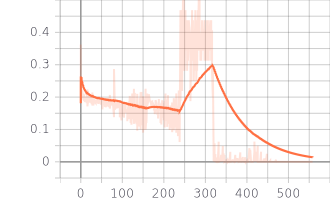}  
   % The printed column width is 8.4 cm.
\caption{{Training loss Vs Iteration with BAGAN}} 
\label{fig:loss Vs iteration}
\endminipage
\end{figure}

Fashion-MNIST is a dataset of Zalando's article images consisting of a training set of 60,000 examples and a test set of 10,000 examples. Each example is a 28x28 grayscale image, associated with a label from 10 classes. (T-shirt/top, Trouser, Pullover, Dress, Coat, Sandal, Shirt, Sneaker, Bag, Ankle boot). We have used all ten classes in the experiment with examples in following ratio: [5000, 4000, 3000, 2000, 1500, 1000, 500, 250, 100, 50]. The data points selected were at random. The true positive and F1 scores of class 0,1 and 9 are 0.90, 0.96, and 0.92, respectively, and the overall F1-score is 0.90. Here in Figure \ref{fig:Loss Vs Iteration} and Figure \ref{fig:loss Vs iteration} shows how training loss varies with iteration for our method compared to BAGAN method. It also shows the time consumed to reach convergence with our model much less than BAGAN and WGAN.

\subsubsection{CIFAR10}
\begin{figure}[ht]
\begin{center}
\includegraphics[width=8cm,height=3cm]{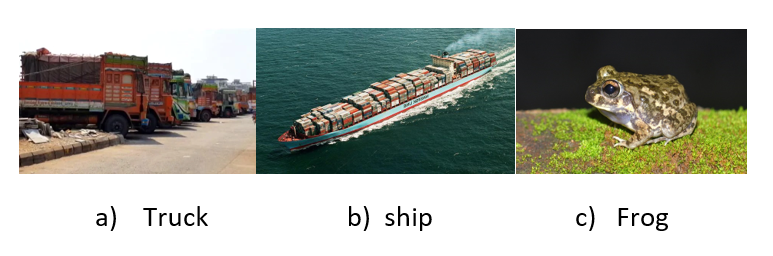}    % The printed column width is 8.4 cm.
\caption{Some images which was earlier miss classified in CIFAR-10 Dataset} 
\label{fig:Results}
\end{center}
\end{figure}

The CIFAR-10 dataset consists of 60000 32x32 color images in 10 classes(airplane, automobile, Bird, Cat, Deer, Dog, Frog, Horse, Ship, Truck) with 6000 images per class. There are 50000 training images and 10000 test images. We have used all ten classes in the experiment with examples in following ratio: [5000, 4500, 4000, 3500, 3000, 2500, 2000, 1500, 1000, 500]. The data points selected were at random. The true positive of class 0,1 and 9 are 0.69, 0.80, and 0.86, respectively, and the overall F1-score of PGAN is 0.75. Some of the image shown in Figure \ref{fig:Results} which were  misclassified earlier with previous methods due to small resolution and now are classified correctly

\begin{figure}[ht]
\begin{center}
\includegraphics[width=5.5cm,height=3cm]{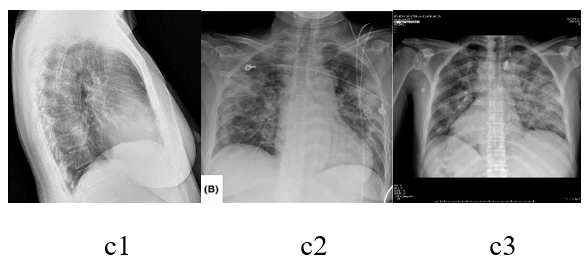}    % The printed column width is 8.4 cm.
\caption{Patient diagnosed from Covid-19, earlier miss classified} 
\label{fig:Results1}
\end{center}
\end{figure}

\subsubsection{ SVHN}
 It can be seen as similar in flavor to MNIST  but incorporates more labeled data. We have used all ten classes in the experiment with examples in following ratio: [5000, 4000, 3000, 2000, 1500, 1000, 500, 250, 100, 50]. The data points selected were at random. The overall F1-score of PGAN is 0.78. 

\subsubsection{ UCF101}
UCF101 is an action recognition data set of real action videos, collected from YouTube, having 101 action categories. It's originally imbalanced classes where "Pushups" have less than 400 samples while "Playing Sitar" has more than 2000 samples. The overall F1-score of PGAN is 0.74.

\subsubsection{ ExDark}
The Exclusively Dark (ExDARK) dataset is a collection of 7,363 low-light images from very low-light environments to twilight (i.e., ten different conditions) with 12 object classes. The overall F1-score of PGAN is 0.54.

\subsubsection{ Covidx}
COVIDx is an open COVID-19 X-Ray dataset comprised of a total of 14200 CXR images. The dataset contains 617 CXR images for COVID-19 cases. For CXR images with no pneumonia and non-COVID19 pneumonia, there are significantly more patient cases and corresponding CXR images. More specifically, there are a total of 8,066 patient cases who have no pneumonia (i.e., regular) and 5,517 patient cases who have non-COVID19 pneumonia. The overall F1-score of PGAN is 0.96.
Some of the images shown in Figure \ref{fig:Results}, which was misclassified earlier with previous methods due to very few samples(0.4\% of total image) classified correctly.

\subsubsection{Kaggle Birds Dataset}
Kaggle Birds dataset\footnote{\url{https://www.kaggle.com/gpiosenka/100-bird-species}} comprises 225 bird species 31316 training images, 1125 test images, and 1125 validation images. We took four species for our experiment: Sora, Pheasant, Road Runner, and Sand Martin. We used the original ratio provided in the dataset, i.e., 300:100:100:100. The overall F1-Score of PGAN is 0.98.

\subsubsection{Satellite Cloud anomaly detection}
Satellite Cloud anomaly detection dataset \footnote{\url{https://www.kaggle.com/ashoksrinivas/cloud-anomaly-detection-images}} consist of two classes cloud and non-cloud, with 1600 images where 100 belong to the cloud, and the rest belongs to non-cloud. We take 50 images for each class for the testing set at random. The overall F1-Score of PGAN is 0.69.
\\
Here in this Table \ref{tab:1}, the first column denotes the various data sets on which we have done experiments, which is also described earlier, And from column 2 to column 10 denotes F-Score with multiple methods. In Table \ref{tab:2} represent the average accuracy of each class on different datasets. Our method outperforms on all datasets.

In our framework, P-GAN acts as an active learning framework using uncertainty sampling. Hence we called the uncertainty approach. While generating labeled data, we make sure that it's just not generate labeled data, but it generates more labeled data where distribution is less.

Our method finds the most useful sample in the given distribution of the least class. There are several ways to measure this. We are using Least Confidence, Margin of Confidence, Ratio of Confidence, and Entropy. The four measures are calculated using the formula given below:
\begin{small}
\begin{align}
    &\text{Least Confidence} = \sum_{i = 0}^{n} \dfrac{m(1 - y_{i}^{*})}{m-1}  \label{eq:10}\\
    &\text{Margin of Confidence} = \sum_{i = 0}^{n}(1 - (y_{i}^{*} - y_{i}^{**})) \label{eq:11} \\
    &\text{Ratio of Confidence} = \sum_{i = 0}^{n} \dfrac{ y_{i}^{**}}{y_{i}^{*}} \label{eq:12} \\
    &\text{Entropy} = - \sum_{i = 0}^{n} \dfrac{y_{i} \log_2{y_{i}}}{\log_2m} \label{eq:13}
\end{align}
\end{small}
% \begin{equation} 
% \begin{split}
% \text{Least Confidence} = \sum_{i = 0}^{n} \dfrac{m(1 - y_{i}^{*})}{m-1}
% \end{split}
% \end{equation}

% \begin{equation} 
% \begin{split}
% \text{Margin of Confidence} = \sum_{i = 0}^{n}(1 - (y_{i}^{*} - y_{i}^{**}))
% \end{split}
% \end{equation}

% \begin{equation} 
% \begin{split}
% \text{Ratio of Confidence} = \sum_{i = 0}^{n} \dfrac{ y_{i}^{**}}{y_{i}^{*}}
% \end{split}
% \end{equation}

% \begin{equation} 
% \begin{split}
% \text{Entropy} = - \sum_{i = 0}^{n} \dfrac{y_{i} \log_2{y_{i}}}{\log_2m}
% \end{split}
% \end{equation}

In the equation \ref{eq:10}, \ref{eq:11}, \ref{eq:12} and \ref{eq:13}, ${y_{i}}$ is the prediction for ${i_{th}}$ sample in the data, ${n}$ represents number of samples in the data, ${m}$ is number of labels in the data. ${y_{i}^{*}}$ and ${y_{i}^{**}}$ are the most confident and second most confident prediction for ${i_{th}}$ sample respectively. Table \ref{tab:3} shows our method generates better samples at the imbalanced class's convex hull boundary.. 

\section{Conclusion}

We have modified the generator network to generate points such that these points lie within the convex hull of training samples. The addition of a classifier, which competes with the generator, helps us generate points at the boundary of classes with better distribution. This proposed method outperforms other state-of-the-art techniques and can be used in computer vision to remove class imbalance. This method can also be used as an uncertainty sampling method (an active learning technique) where data annotation is challenging. Our future work will be to test it on NLP(Click Through Rate, Anamoly Detection, Semantic segmentation, and others) and Speech tasks(with less annotated low resource language). We would also like to use different generator discriminator architecture such as Memory Replay GAN.

{\small
\bibliographystyle{ieee_fullname}
\bibliography{ref}

\begin{thebibliography}{10}\itemsep=-1pt

\bibitem{ali2015classification}
Aida Ali, Siti~Mariyam Shamsuddin, Anca~L Ralescu, et~al.
\newblock Classification with class imbalance problem: a review.
\newblock {\em Int. J. Advance Soft Compu. Appl}, 7(3):176--204, 2015.

\bibitem{arjovsky2017wasserstein}
Martin Arjovsky, Soumith Chintala, and L{\'e}on Bottou.
\newblock Wasserstein gan.
\newblock {\em arXiv preprint arXiv:1701.07875}, 2017.

\bibitem{bergmann2019mvtec}
Paul Bergmann, Michael Fauser, David Sattlegger, and Carsten Steger.
\newblock Mvtec ad--a comprehensive real-world dataset for unsupervised anomaly
  detection.
\newblock In {\em Proceedings of the IEEE Conference on Computer Vision and
  Pattern Recognition}, pages 9592--9600, 2019.

\bibitem{chau2011border}
Asdr{\'u}bal~L{\'o}pez Chau, Xiaoou Li, Wen Yu, Jair Cervantes, and Pedro
  Mej{\'\i}a-{\'A}lvarez.
\newblock Border samples detection for data mining applications using non
  convex hulls.
\newblock In {\em Mexican International Conference on Artificial Intelligence},
  pages 261--272. Springer, 2011.

\bibitem{chawla2002smote}
Nitesh~V Chawla, Kevin~W Bowyer, Lawrence~O Hall, and W~Philip Kegelmeyer.
\newblock Smote: synthetic minority over-sampling technique.
\newblock {\em Journal of artificial intelligence research}, 16:321--357, 2002.

\bibitem{cohen2020covid}
Joseph~Paul Cohen, Paul Morrison, Lan Dao, Karsten Roth, Tim~Q Duong, and
  Marzyeh Ghassemi.
\newblock Covid-19 image data collection: Prospective predictions are the
  future.
\newblock {\em arXiv preprint arXiv:2006.11988}, 2020.

\bibitem{douzas2018effective}
Georgios Douzas and Fernando Bacao.
\newblock Effective data generation for imbalanced learning using conditional
  generative adversarial networks.
\newblock {\em Expert Systems with applications}, 91:464--471, 2018.

\bibitem{gauthier2014conditional}
Jon Gauthier.
\newblock Conditional generative adversarial nets for convolutional face
  generation.
\newblock {\em Class Project for Stanford CS231N: Convolutional Neural Networks
  for Visual Recognition, Winter semester}, 2014(5):2, 2014.

\bibitem{george2015image}
Marian George.
\newblock Image parsing with a wide range of classes and scene-level context.
\newblock In {\em Proceedings of the IEEE conference on computer vision and
  pattern recognition}, pages 3622--3630, 2015.

\bibitem{goodfellow2014generative}
Ian Goodfellow, Jean Pouget-Abadie, Mehdi Mirza, Bing Xu, David Warde-Farley,
  Sherjil Ozair, Aaron Courville, and Yoshua Bengio.
\newblock Generative adversarial nets.
\newblock In {\em Advances in neural information processing systems}, pages
  2672--2680, 2014.

\bibitem{goodfellow2013multi}
Ian~J Goodfellow, Yaroslav Bulatov, Julian Ibarz, Sacha Arnoud, and Vinay Shet.
\newblock Multi-digit number recognition from street view imagery using deep
  convolutional neural networks.
\newblock {\em arXiv preprint arXiv:1312.6082}, 2013.

\bibitem{huang2016learning}
Chen Huang, Yining Li, Chen~Change Loy, and Xiaoou Tang.
\newblock Learning deep representation for imbalanced classification.
\newblock In {\em Proceedings of the IEEE conference on computer vision and
  pattern recognition}, pages 5375--5384, 2016.

\bibitem{japkowicz2002class}
Nathalie Japkowicz and Shaju Stephen.
\newblock The class imbalance problem: A systematic study.
\newblock {\em Intelligent data analysis}, 6(5):429--449, 2002.

\bibitem{johnson2019survey}
Justin~M Johnson and Taghi~M Khoshgoftaar.
\newblock Survey on deep learning with class imbalance.
\newblock {\em Journal of Big Data}, 6(1):27, 2019.

\bibitem{khan2017cost}
Salman~H Khan, Munawar Hayat, Mohammed Bennamoun, Ferdous~A Sohel, and Roberto
  Togneri.
\newblock Cost-sensitive learning of deep feature representations from
  imbalanced data.
\newblock {\em IEEE transactions on neural networks and learning systems},
  29(8):3573--3587, 2017.

\bibitem{kotsiantis2006handling}
Sotiris Kotsiantis, Dimitris Kanellopoulos, Panayiotis Pintelas, et~al.
\newblock Handling imbalanced datasets: A review.
\newblock {\em GESTS International Transactions on Computer Science and
  Engineering}, 30(1):25--36, 2006.

\bibitem{krizhevsky2009learning}
Alex Krizhevsky, Geoffrey Hinton, et~al.
\newblock Learning multiple layers of features from tiny images.
\newblock 2009.

\bibitem{lecun1998mnist}
Yann LeCun.
\newblock The mnist database of handwritten digits.
\newblock {\em http://yann. lecun. com/exdb/mnist/}, 1998.

\bibitem{ledig2017photo}
Christian Ledig, Lucas Theis, Ferenc Husz{\'a}r, Jose Caballero, Andrew
  Cunningham, Alejandro Acosta, Andrew Aitken, Alykhan Tejani, Johannes Totz,
  Zehan Wang, et~al.
\newblock Photo-realistic single image super-resolution using a generative
  adversarial network.
\newblock In {\em Proceedings of the IEEE conference on computer vision and
  pattern recognition}, pages 4681--4690, 2017.

\bibitem{lee2012anomaly}
Yuh-Jye Lee, Yi-Ren Yeh, and Yu-Chiang~Frank Wang.
\newblock Anomaly detection via online oversampling principal component
  analysis.
\newblock {\em IEEE transactions on knowledge and data engineering},
  25(7):1460--1470, 2012.

\bibitem{li2018cost}
Zhiqiang Li, Xiao-Yuan Jing, Fei Wu, Xiaoke Zhu, Baowen Xu, and Shi Ying.
\newblock Cost-sensitive transfer kernel canonical correlation analysis for
  heterogeneous defect prediction.
\newblock {\em Automated Software Engineering}, 25(2):201--245, 2018.

\bibitem{loh2019getting}
Yuen~Peng Loh and Chee~Seng Chan.
\newblock Getting to know low-light images with the exclusively dark dataset.
\newblock {\em Computer Vision and Image Understanding}, 178:30--42, 2019.

\bibitem{lu2015feature}
Xiaoyong Lu, Musheng Chen, Jhenglong Wu, and Peichan Chan.
\newblock A feature-partition and under-sampling based ensemble classifier for
  web spam detection.
\newblock {\em International Journal of Machine Learning and Computing},
  5(6):454, 2015.

\bibitem{mariani2018bagan}
Giovanni Mariani, Florian Scheidegger, Roxana Istrate, Costas Bekas, and
  Cristiano Malossi.
\newblock Bagan: Data augmentation with balancing gan.
\newblock {\em arXiv preprint arXiv:1803.09655}, 2018.

\bibitem{mathew2017classification}
Josey Mathew, Chee~Khiang Pang, Ming Luo, and Weng~Hoe Leong.
\newblock Classification of imbalanced data by oversampling in kernel space of
  support vector machines.
\newblock {\em IEEE transactions on neural networks and learning systems},
  29(9):4065--4076, 2017.

\bibitem{menendez1997jensen}
ML Men{\'e}ndez, JA Pardo, L Pardo, and MC Pardo.
\newblock The jensen-shannon divergence.
\newblock {\em Journal of the Franklin Institute}, 334(2):307--318, 1997.

\bibitem{mirza2014conditional}
Mehdi Mirza and Simon Osindero.
\newblock Conditional generative adversarial nets.
\newblock {\em arXiv preprint arXiv:1411.1784}, 2014.

\bibitem{mostajabi2015feedforward}
Mohammadreza Mostajabi, Payman Yadollahpour, and Gregory Shakhnarovich.
\newblock Feedforward semantic segmentation with zoom-out features.
\newblock In {\em Proceedings of the IEEE conference on computer vision and
  pattern recognition}, pages 3376--3385, 2015.

\bibitem{netzer2019street}
Yuval Netzer, Tao Wang, Adam Coates, Alessandro Bissacco, Bo Wu, and A Ng.
\newblock The street view house numbers (svhn) dataset, 2019.

\bibitem{nguyen2009learning}
Giang~Hoang Nguyen, Abdesselam Bouzerdoum, and Son~Lam Phung.
\newblock Learning pattern classification tasks with imbalanced data sets.
\newblock {\em Pattern recognition}, pages 193--208, 2009.

\bibitem{odena2017conditional}
Augustus Odena, Christopher Olah, and Jonathon Shlens.
\newblock Conditional image synthesis with auxiliary classifier gans.
\newblock In {\em International conference on machine learning}, pages
  2642--2651, 2017.

\bibitem{ramentol2012smote}
Enislay Ramentol, Yail{\'e} Caballero, Rafael Bello, and Francisco Herrera.
\newblock Smote-rsb*: a hybrid preprocessing approach based on oversampling and
  undersampling for high imbalanced data-sets using smote and rough sets
  theory.
\newblock {\em Knowledge and information systems}, 33(2):245--265, 2012.

\bibitem{schmidhuber2020generative}
J{\"u}rgen Schmidhuber.
\newblock Generative adversarial networks are special cases of artificial
  curiosity (1990) and also closely related to predictability minimization
  (1991).
\newblock {\em Neural Networks}, 2020.

\bibitem{sharma2017evidence}
Manali Sharma and Mustafa Bilgic.
\newblock Evidence-based uncertainty sampling for active learning.
\newblock {\em Data Mining and Knowledge Discovery}, 31(1):164--202, 2017.

\bibitem{ganmanga}
Nikita Sharma.
\newblock My mangagan: Building my first generative adversarial network.

\bibitem{snoek2012practical}
Jasper Snoek, Hugo Larochelle, and Ryan~P Adams.
\newblock Practical bayesian optimization of machine learning algorithms.
\newblock In {\em Advances in neural information processing systems}, pages
  2951--2959, 2012.

\bibitem{soomro2012ucf101}
Khurram Soomro, Amir~Roshan Zamir, and Mubarak Shah.
\newblock Ucf101: A dataset of 101 human actions classes from videos in the
  wild.
\newblock {\em arXiv preprint arXiv:1212.0402}, 2012.

\bibitem{springenberg2015unsupervised}
Jost~Tobias Springenberg.
\newblock Unsupervised and semi-supervised learning with categorical generative
  adversarial networks.
\newblock {\em arXiv preprint arXiv:1511.06390}, 2015.

\bibitem{srivastava2014dropout}
Nitish Srivastava, Geoffrey Hinton, Alex Krizhevsky, Ilya Sutskever, and Ruslan
  Salakhutdinov.
\newblock Dropout: a simple way to prevent neural networks from overfitting.
\newblock {\em The journal of machine learning research}, 15(1):1929--1958,
  2014.

\bibitem{suh2019generative}
Sungho Suh, Haebom Lee, Jun Jo, Paul Lukowicz, and Yong~Oh Lee.
\newblock Generative oversampling method for imbalanced data on bearing fault
  detection and diagnosis.
\newblock {\em Applied Sciences}, 9(4):746, 2019.

\bibitem{suh2020cegan}
Sungho Suh, Haebom Lee, Paul Lukowicz, and Yong~Oh Lee.
\newblock Cegan: Classification enhancement generative adversarial networks for
  unraveling data imbalance problems.
\newblock {\em Neural Networks}, 2020.

\bibitem{thai2010cost}
Nguyen Thai-Nghe, Zeno Gantner, and Lars Schmidt-Thieme.
\newblock Cost-sensitive learning methods for imbalanced data.
\newblock In {\em The 2010 International joint conference on neural networks
  (IJCNN)}, pages 1--8. IEEE, 2010.

\bibitem{vuttipittayamongkol2020improved}
Pattaramon Vuttipittayamongkol and Eyad Elyan.
\newblock Improved overlap-based undersampling for imbalanced dataset
  classification with application to epilepsy and parkinson’s disease.
\newblock {\em International journal of neural systems}, 30(08):2050043, 2020.

\bibitem{wang2020covid}
Linda Wang and Alexander Wong.
\newblock Covid-net: A tailored deep convolutional neural network design for
  detection of covid-19 cases from chest x-ray images.
\newblock {\em arXiv preprint arXiv:2003.09871}, 2020.

\bibitem{wei2015deep}
Xiu-Shen Wei, Bin-Bin Gao, and Jianxin Wu.
\newblock Deep spatial pyramid ensemble for cultural event recognition.
\newblock In {\em Proceedings of the IEEE international conference on computer
  vision workshops}, pages 38--44, 2015.

\bibitem{xiao2017fashion}
Han Xiao, Kashif Rasul, and Roland Vollgraf.
\newblock Fashion-mnist: a novel image dataset for benchmarking machine
  learning algorithms.
\newblock {\em arXiv preprint arXiv:1708.07747}, 2017.

\bibitem{yan2015deep}
Yilin Yan, Min Chen, Mei-Ling Shyu, and Shu-Ching Chen.
\newblock Deep learning for imbalanced multimedia data classification.
\newblock In {\em 2015 IEEE international symposium on multimedia (ISM)}, pages
  483--488. IEEE, 2015.

\bibitem{yang2017lr}
Jianwei Yang, Anitha Kannan, Dhruv Batra, and Devi Parikh.
\newblock Lr-gan: Layered recursive generative adversarial networks for image
  generation.
\newblock {\em arXiv preprint arXiv:1703.01560}, 2017.

\bibitem{yang2015multi}
Yi Yang, Zhigang Ma, Feiping Nie, Xiaojun Chang, and Alexander~G Hauptmann.
\newblock Multi-class active learning by uncertainty sampling with diversity
  maximization.
\newblock {\em International Journal of Computer Vision}, 113(2):113--127,
  2015.

\bibitem{yen2006under}
Show-Jane Yen and Yue-Shi Lee.
\newblock Under-sampling approaches for improving prediction of the minority
  class in an imbalanced dataset.
\newblock In {\em Intelligent Control and Automation}, pages 731--740.
  Springer, 2006.

\end{thebibliography}
}

\newpage
\onecolumn
\appendix
\section{Appendix}
\subsection{CGAN}
Generative adversarial nets can be extended to a conditional model if both the generator and discriminator are conditioned on some extra information y. y could be any kind of auxiliary information, such as class labels or data from other modalities. In CGAN, the conditioning is performed by feeding y into the both the discriminator and generator as additional input layer. The objective function for CGAN is similar to GAN with only difference of conditioning by class label ${y}$:
\begin{figure}[ht]
\begin{center}
\includegraphics[width=10.0cm]{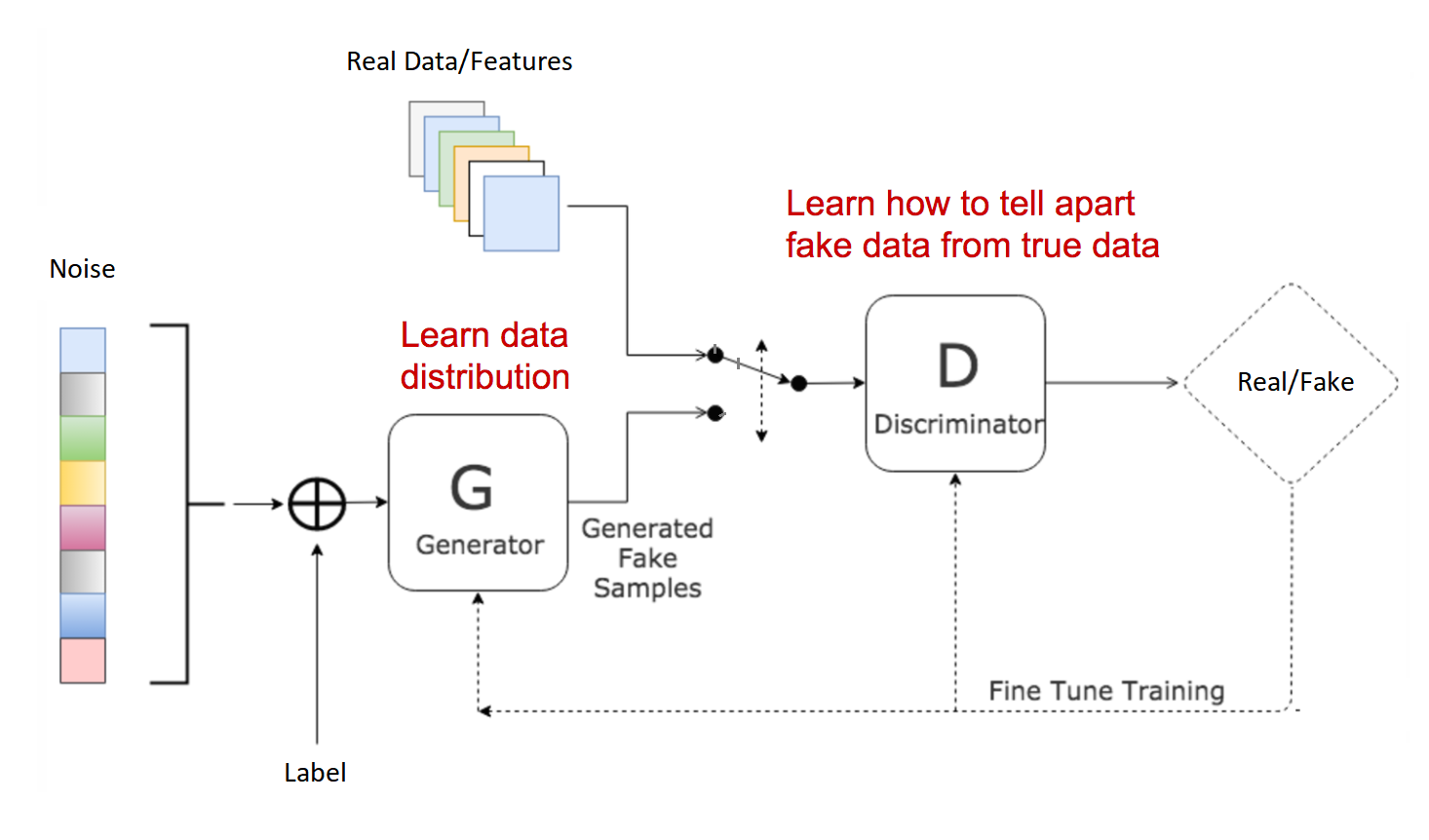}  
\caption{{Framework of Conditional GAN (CGAN).}} 
\label{fig:CGAN}
\end{center}
\end{figure}

\begin{equation} \label{eq:2}
\begin{split}
\min_{G} \max_{D} V(G,D) =  \mathbb{E}_{x \sim p_{\text{data}}(x)}[\log D(x|y)]   + \mathbb{E}_{z \sim p_{z}(z)}[\log (1 - D(G(z|y)))]
\end{split}
\end{equation}

\subsection{AC-GAN}
The Auxiliary Classifier GAN, or AC-GAN for short, is an extension of the conditional GAN that changes the discriminator to predict the class label of a given image rather than receive it as input. It has the effect of stabilizing the training process and allowing the generation of large high-quality images whilst learning a representation in the latent space that is independent of the class label. Though it still suffers in training due to contradicting objective function. The objective function ACGAN is the summation of two function below: 
\begin{figure}[ht]
\begin{center}
\includegraphics[width=10.0cm]{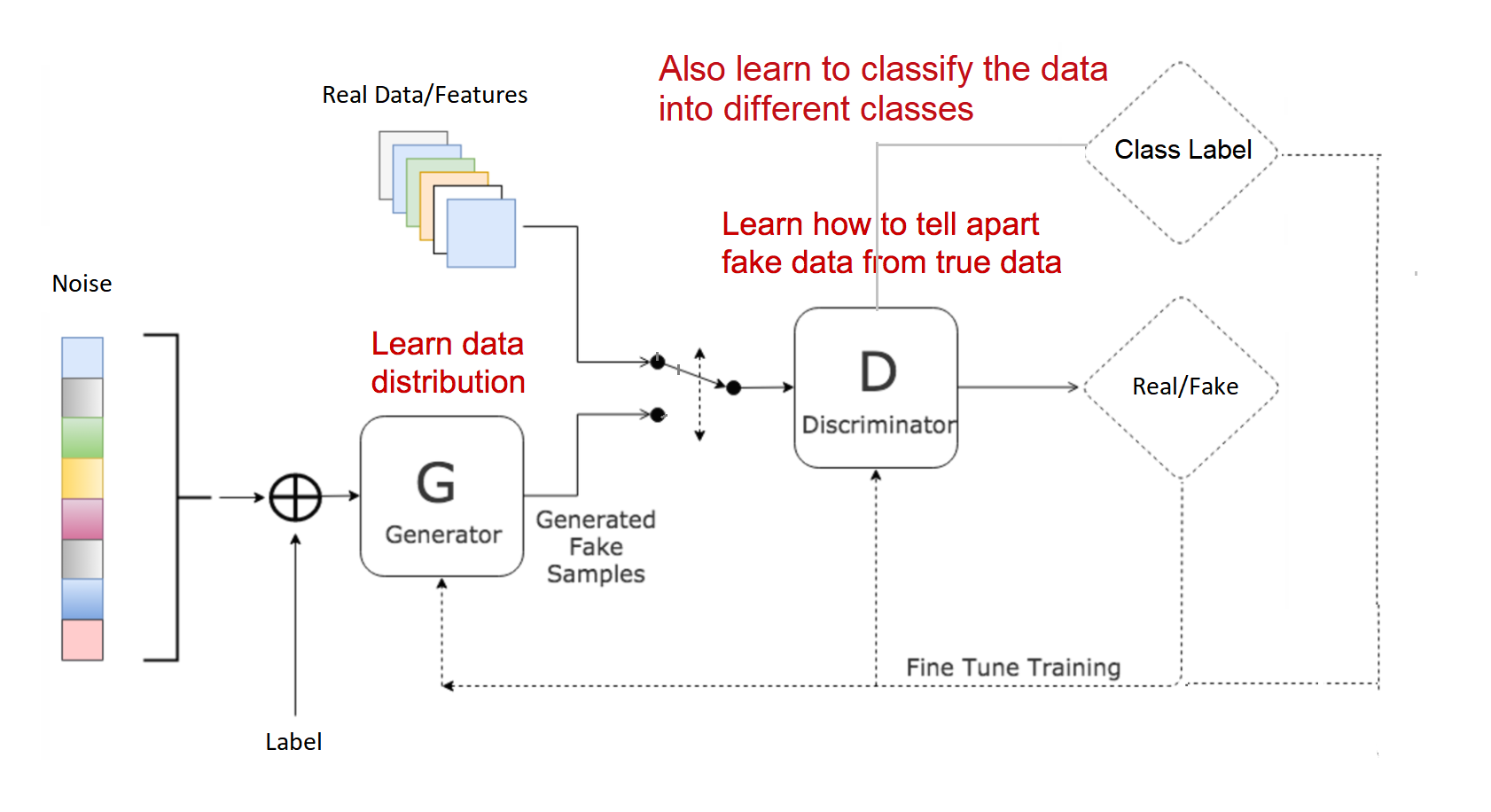}    % The printed column width is 8.4 cm.
\caption{{Framework of Auxiliary-Classifier GAN (AC-GAN).}} 
\label{fig:ACGAN}
\end{center}
\end{figure}

 \begin{equation} \label{eq:3}
\begin{split}
L_{S} &= \mathbb{E}[\log P(S = real | X_{real})]  +\mathbb{E}[\log P(S = fake | X_{fake})]
\end{split}
\end{equation}

 \begin{equation} \label{eq:4}
\begin{split}
L_{C} &= \mathbb{E}[\log P(C = c | X_{real})]  + \mathbb{E}[\log P(C = c | X_{fake})]
\end{split}
\end{equation}

In equation \ref{eq:3} and \ref{eq:4}, $P(S|X)$ is the probability distribution over source and $P(C|X)$ is the probability distribution over the class labels.

\subsection{KL Divergence}
KL (Kullback–Leibler) Divergence measures how one probability distribution ${p}$ diverges from a second expected probability distribution ${q}$.

\begin{equation} \label{eq:14}
\begin{split}
D_{KL}(p||q) = \int_{x}^{}p(x)\log \dfrac{p(x)}{q(x)}dx
\end{split}
\end{equation}

${D_{KL}}$ achieves the minimum zero when ${p(x) == q(x)}$ everywhere. It is noticeable according to the formula that KL divergence is asymmetric. In cases where ${p(x)}$ is close to zero, but ${q(x)}$ is significantly non-zero, the q’s effect is disregarded. It could cause buggy results when we just want to measure the similarity between two equally important distributions.

\subsection{Wasserstein Distance}
Wasserstein Distance is a measure of the distance between two probability distributions. It is also called Earth Mover’s distance, short for EM distance, because informally it can be interpreted as the minimum energy cost of moving and transforming a pile of dirt in the shape of one probability distribution to the shape of the other distribution. The cost is quantified by: the amount of dirt moved x the moving distance. When dealing with the continuous probability domain, the distance formula becomes:

\begin{equation} \label{eq:15}
\begin{split}
W(p_{r}, p_{g}) = \inf_{\gamma \sim \Pi(p_{r}, p_{g})} \mathbb{E}_{(x,y) \sim \gamma} [\|x-y\|]
\end{split}
\end{equation}

In the formula above, ${\Pi(p_{r}, p_{g})}$ is the set of all possible joint probability distributions between ${p_{r}}$ and ${p_{s}}$. One joint distribution ${\gamma \in \Pi(p_{r}, p_{g})}$ describes one dirt transport plan, same as the discrete example above, but in the continuous probability space. Precisely ${\gamma (x,y)}$ states the percentage of dirt should be transported from point ${x}$ to ${y}$ so as to make ${x}$ follows the same probability distribution of ${y}$. That’s why the marginal distribution over ${x}$ adds up to ${p_{g}}$, ${\sum_{x}\gamma(x,y) = p_{g}(y)}$ (Once we finish
moving the planned amount of dirt from every possible ${x}$ to the target ${y}$, we end up with exactly what ${y}$ has according to ${p_{g}}$.) and vice versa ${\sum_{y}\gamma(x,y) = p_{r}(x)}$. When treating ${x}$ as the starting point and ${y}$ as the destination, the total amount of dirt moved is ${\gamma (x,y)}$ and the traveling distance is $\|x-y\|$ and thus the cost is ${\gamma (x,y).\|x-y\|}$. The expected cost averaged across all the ${(x, y)}$ pairs can be easily computed as:

\begin{equation} \label{eq:16}
\begin{split}
\sum_{x,y}\gamma (x,y).\|x-y\| = \mathbb{E}_{x,y \sim \gamma} \|x-y\|
\end{split}
\end{equation}

Finally, we take the minimum one among the costs of all dirt moving solutions as the EM distance. In the definition of Wasserstein distance, the inf (infimum, also known as *greatest lower bound*) indicates that we are only interested in the smallest cost. For more details you can refer the paper \footnote{\url{https://arxiv.org/pdf/1904.08994.pdf}}.

\end{document}